%% file: paper.tex
\documentclass[conference]{IEEEtran}
\IEEEoverridecommandlockouts
\usepackage[font=small,labelfont=bf]{caption}

\usepackage[compact]{titlesec}
\titlespacing\section{0pt}{5pt plus 4pt minus 2pt}{0pt plus 2pt minus 2pt}
\titlespacing\subsection{0pt}{1pt plus 1pt minus 2pt}{0pt plus 1pt minus 2pt}
\titlespacing\subsubsection{0pt}{2pt plus 4pt minus 2pt}{0pt plus 2pt minus 2pt}

\newcommand{\ie}{\textit{i}.\textit{e}.,}
\newcommand{\eg}{\textit{e}.\textit{g}.,}

\usepackage{float}
\usepackage{enumitem}

\usepackage{cite}
\usepackage{mathtools, braket}
\usepackage{amsmath,amssymb,amsfonts}
\usepackage{algorithm}
\usepackage{algpseudocode}
\usepackage{graphicx}
\usepackage{textcomp}
\usepackage{xcolor}
\usepackage{wrapfig}
\usepackage{multicol}
\usepackage{tabularray}
\usepackage{multirow}
\usepackage{lipsum}
\usepackage{svg}
\usepackage{subcaption}
\usepackage{caption}
\usepackage[normalem]{ulem}
\usepackage{tabularx,booktabs,ragged2e}
\newcolumntype{L}{>{\RaggedRight\arraybackslash}X}
\usepackage{diagbox}
\usepackage{pifont}
\usepackage{cleveref}
\newcommand{\cmark}{\ding{51}}
\newcommand{\xmark}{\ding{55}}
\definecolor{limeyellow}{RGB}{228,255,100}
\usepackage[colorinlistoftodos,prependcaption,
textsize=footnotesize,
color=limeyellow,linecolor=magenta]{todonotes}
\presetkeys{todonotes}{inline}{}

\makeatletter
\def\algbackskip{\hskip-\ALG@thistlm}
\makeatother

\usepackage{subfiles} 

\usepackage[a4paper, total={184mm,239mm}]{geometry}
\def\BibTeX{{\rm B\kern-.05em{\sc i\kern-.025em b}\kern-.08em
    T\kern-.1667em\lower.7ex\hbox{E}\kern-.125emX}}
\begin{document}

\linepenalty=1000

\title{
Context-aware Multi-Model Object Detection for Diversely Heterogeneous Compute Systems
}

\newcommand{\sysname}[1]{\textit{SHIFT}}

\IEEEaftertitletext{\vspace{-2.0\baselineskip}}

\author{\IEEEauthorblockN{Justin Davis}
\IEEEauthorblockA{\textit{Computer Science} \\
\textit{Colorado School of Mines}\\
jcdavis@mines.edu}
\and
\IEEEauthorblockN{Mehmet E. Belviranli}
\IEEEauthorblockA{\textit{Computer Science} \\
\textit{Colorado School of Mines}\\
belviranli@mines.edu}
}

\maketitle

\subfile{00-abstract.tex}

\begin{IEEEkeywords}
accelerator, autonomous, context-aware, object detection, gpu, heterogeneous
\end{IEEEkeywords}

\subfile{01-introduction.tex}
\subfile{02-relatedworks.tex}

\subfile{03-methodology.tex}
\subfile{04-experiments.tex}
\subfile{05-results.tex}
\subfile{06-conclusion.tex}

\section*{Acknowledgements}
This material is based upon work supported by the National Science Foundation (NSF) under Grants No. CCF-2124010 and CHE-2235143. Any opinions, findings, or recommendations expressed in this material are those of the authors and do not necessarily reflect the views of NSF.

\bibliographystyle{IEEEtran}
\bibliography{bib}

\end{document}

%% file: 00-abstract.tex
\begin{abstract}

In recent years, deep neural networks (DNNs) have gained widespread adoption for continuous mobile object detection (OD) tasks, particularly in autonomous systems. However, a prevalent issue in their deployment is the one-size-fits-all approach, where a single DNN is used, resulting in inefficient utilization of computational resources. This inefficiency is particularly detrimental in energy-constrained systems, as it degrades overall system efficiency. We identify that, 
the contextual information embedded in the input data stream (\eg{} the frames in the camera feed that the OD models are run on) could be exploited to allow a more efficient multi-model-based OD process. In this paper, we propose \sysname{} which continuously selects from a variety of DNN-based OD models depending on the dynamically changing contextual information and computational constraints. During this selection, \sysname{} uniquely considers multi-accelerator execution to better optimize the energy-efficiency while satisfying the latency constraints. Our proposed methodology results in improvements of up to \textbf{7.5x} in energy usage and \textbf{2.8x} in latency compared to state-of-the-art GPU-based single model OD approaches. 

\end{abstract}

%% file: 01-introduction.tex
\section{Introduction}\label{chap:intro}

Modern autonomous systems employ deep neural networks (DNN) for various tasks and all-in-one system-on-chips (SoC) to enable decision making on-the-go without human intervention. Typically, such autonomous systems are equipped with SoCs featuring graphical processing units (GPU) for the execution of DNNs. A common and critical autonomous task is object detection (OD) which identifies objects of interest in the environment, captured by the stream of images obtained by the camera. A common practice employed by system developers is to select and configure a single DNN, such as YoloV7\cite{yolov7},
and map it to the fastest processor in the SoC, which is typically a GPU.
In this conventional setup, there is limited room for improving the latency and/or energy usage of the autonomous system, as the model and the target processing unit is fixed.
In response, several studies~\cite{glimpse, flexpatch} propose offloading the computation to a remote server, while others~\cite{adavp, marlin, framehopper} attempt to reduce the computational demand by modifying the underlying model or using a subset of the data stream. However, offloading is not a viable option due to the latency overhead associated with remote processing. 
On the other hand, modifying models or selectively skipping data
often results in a significant compromise in accuracy. Instead, in this work, we explore optimizing the system performance by employing a context-aware multi-model execution and leveraging different type of accelerators available in SoCs. \looseness=-1

\begin{figure}
\begin{center}
    \includegraphics[scale=0.21]{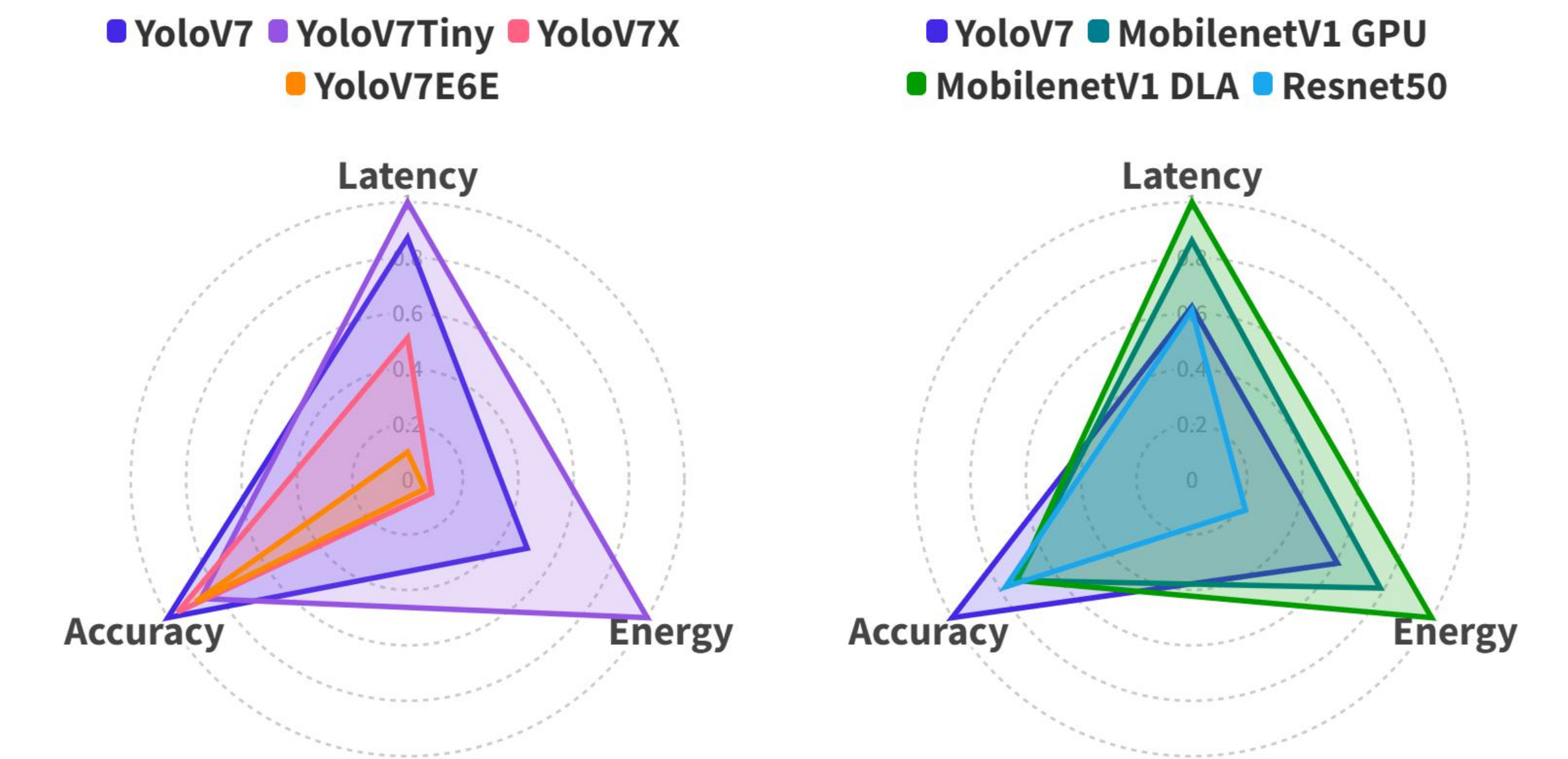}
    \vspace{-1.5em}
    \caption{Comparison of (a) single-model with multiple parameter sizes on the left against (b) multi-model object detection architectures on the right. The larger the value along each axis the better: a perfect model would be largest triangle across all axes.}
    \label{fig:pyramid}
\end{center}
\vspace{-2.5em}
\end{figure}

\begin{figure}[b]
\vspace{-1.5em}
\begin{center}
    \includegraphics[scale=0.023]{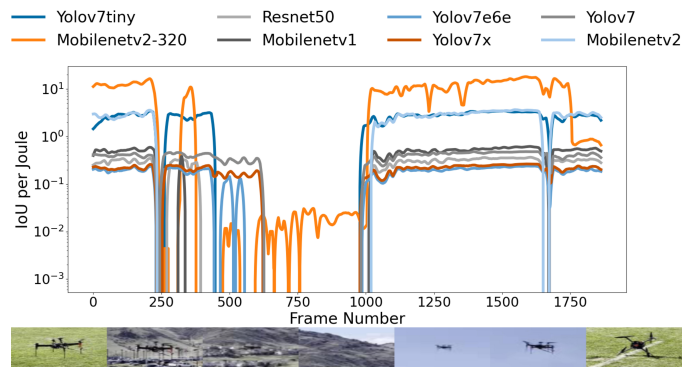}
    \caption{Single model object detection efficiency on GPU for commonly used DNNs and their variations on a test set for continuous detection and tracking of an aerial drone. Efficiency is quantified by intersection over union (IoU) per Joule of energy (see Sec.~\ref{chap:experiments} for details). }
    \label{fig:showcase}
\vspace{-0.8em}
\end{center}
\end{figure}

Modern SoCs often embed neural network (NN) accelerators alongside with GPUs to perform  low-power DNN inference. For example, the Nvidia DLA
in the Jetson Xavier series
allows DNN inference with up to 2.5x energy savings
compared to the GPU in the same SoC, at the cost of 2x slower latency. 
Another accelerator, the RCV2 on the OAK-D from Luxonis, provides 4 TOPs of computation at 5 Watts, surpassing GPU energy efficiency.
Having different types of DNN accelerators in the system enables a tunable trade-off between latency and energy~\cite{axonn}. \looseness=-1

A common approach to save energy and reduce computational demands in the OD process is to quantize a given object detection model (ODM) and create less accurate but more energy and performance efficient versions of the same model \cite{gholami2021survey}. While this method allows for the execution of state-of-the-art models on performance-limited SoCs, efficiency-related optimizations are constrained due to the entire OD process being confined to a single DNN model. A more comprehensive and flexible way to further increase energy savings is to dynamically switch to simpler, less compute demanding ODMs if their detection accuracy is sufficient for the changing (\ie{} \textit{contextual}) characteristics of the captured frames. For example, if the target object being detected is in front of a solid, contrasted background within a close distance, then both simple and advanced models perform equally well. Fig.~\ref{fig:showcase} demonstrates this \textit{contextual} change by depicting the timeline of varying accuracies that different ODMs exhibit on one of our test sets (see Sec.~\ref{chap:results} for details). Through optimal utilization of all available heterogeneous models, we observed enhancements in accuracy by 3\%, latency by 5.2x, and energy usage by 13.6x, or a combination thereof.
We illustrate the three-way energy-accuracy-latency (\textit{e-a-l}) relationship between the standard YOLOv7 model~\cite{yolov7} and other simpler model families in Fig.~\ref{fig:pyramid}.
While smaller variations of YOLOv7  (Fig.~\ref{fig:pyramid}.a) result in a monotonic decrease of energy and latency; the use of different type of ODMs (Fig.~\ref{fig:pyramid}.b) exhibits a non-monotonically changing relationship between the three metrics. Detailed statistics on the \textit{e-a-l} trade-off for different accelerators and the CPU could be found in Table~\ref{table:initial-hetero-experiment}.\looseness=-1

\begin{table}[t]
    \setlength\tabcolsep{2.4pt}
    \begin{tabularx}{\columnwidth}{l|c|ccc|ccc|ccc}
        \toprule
        \multirow{1}{*}{Model} & \multicolumn{1}{c}{IoU} & \multicolumn{3}{c}{Inference (s)} & \multicolumn{3}{c}{Power (W)} & \multicolumn{3}{c}{Energy (J)} \\
        &  & CPU & GPU & DLA & CPU & GPU & DLA & CPU & GPU & DLA \\
        \midrule
          YoloV7 & 0.62 & 1.65 & 0.13 & 0.12 & 7.60 & 15.1 & 15.1 & 20.5 & 1.97 & 1.78\\
          YoloV7Tiny & 0.53 & 0.38 & 0.03 & 0.02 & 7.20 & 11.2 & 11.2 & 4.19 & 0.28 & 0.27\\
          MobilenetV1 & 0.45 & - & 0.09 & 0.09 & - & 16.2 & 6.10 & - & 1.52 & 0.56\\
        \bottomrule
    \end{tabularx}
    \caption{Average statistics for two architectures of object detection models and their performance on CPU, GPU, and GPU/DLA.
    }
    \label{table:initial-hetero-experiment}
    \vspace{-2em}
\end{table}

\textit{By leveraging the optimization potential offered by multi-accelerator systems and multi-model object detection, there exists an opportunity to enhance accuracy, reduce latency, and conserve energy in real-time operations within autonomous systems.}
However, creating such an OD scheme presents several challenges:
(i) Models need to be characterized in advance to determine their performance and energy characteristics. (ii) Since the accuracy of each model depends on the dynamic context, predicting accuracies without running them for every encountered frame is not trivial. 
(iii) Not all models considered by the system can be simultaneously loaded into memory due to limitations in available resources.

In this paper, we present \sysname{} to enable energy and latency efficient multi-model and multi-accelerator object detection for autonomous systems. This approach takes into consideration the changing contextual features of the input frames, accelerators in the system, and ODMs with different execution characteristics. Our contributions are as follows:
\begin{itemize}[leftmargin=*]
    \item We build a unique graph-based mechanism, named confidence graphs, that is used to rapidly predict the changing accuracy of different ODMs during runtime.
    \item We create a novel scheduler that can adapt to specific system constraints by targeting model accuracy, latency, or energy consumption, based on real-time contextual information derived from the input stream of frames. This scheduler decides the ODM and accelerator to run such that system objectives are achieved under given constraints.
    \item We deploy a dynamic model loading mechanism that is capable of managing memory resources for each ODM and facilitates switching between ODMs when necessary.
    \item We evaluate the utility and efficiency of \sysname{} on three unique off-the-shelf accelerators designed for autonomous systems. We show that our proposed methodology results in improvements of up to \textbf{7.5x} in energy usage and \textbf{2.8x} in latency compared to GPU-based single model OD approaches.
\end{itemize}

%% file: 02-relatedworks.tex
\section{Related Work}\label{chap:related-work}

\begin{table}[t]
\scriptsize
\centering
\begin{tabular}{ | m{2.3cm}|  m{0.3cm} | m{0.3cm}| m{0.3cm}| m{0.3cm}| m{0.3cm} | m{0.3cm} | m{0.3cm} |  m{0.3cm} |  }  
  \hline
  \diagbox[width=2.7cm, height=1cm, innerleftsep=0.25cm]{Feature}{Related Work}
  & \rotatebox{90}{Glimpse~\cite{glimpse}}
  & \rotatebox{90}{MARLIN~\cite{marlin}}
  & \rotatebox{90}{AdaVP~\cite{adavp}}
  & \rotatebox{90}{RoaD-RuNNer~\cite{roadrunner}}
  & \rotatebox{90}{Fast UQ~\cite{nvidia-pose}}
  & \rotatebox{90}{Herald~\cite{herald}}
  & \rotatebox{90}{AxoNN~\cite{axonn}}
  & \rotatebox{90}{\textbf{\sysname{}}}
   \\ \hline 
  \begin{tabular}{@{}c@{}} Context Awareness\end{tabular} 
  & \multicolumn{1}{c|}\xmark
  & \multicolumn{1}{c|}\cmark
  & \multicolumn{1}{c|}\cmark
  & \multicolumn{1}{c|}\cmark
  & \multicolumn{1}{c|}\xmark
  & \multicolumn{1}{c|}\xmark
  & \multicolumn{1}{c|}\xmark
  & \multicolumn{1}{c|}\cmark
  \\ \hline
  \begin{tabular}{@{}c@{}} Multi-Accelerator\end{tabular}
  & \multicolumn{1}{c|}\xmark
  & \multicolumn{1}{c|}\xmark
  & \multicolumn{1}{c|}\xmark
  & \multicolumn{1}{c|}\xmark
  & \multicolumn{1}{c|}\xmark
  & \multicolumn{1}{c|}\cmark
  & \multicolumn{1}{c|}\cmark
  & \multicolumn{1}{c|}\cmark
  \\ \hline
  \begin{tabular}{@{}l@{}} Multi-DNN\end{tabular}
  & \multicolumn{1}{c|}\xmark
  & \multicolumn{1}{c|}\xmark
  & \multicolumn{1}{c|}\xmark
  & \multicolumn{1}{c|}\xmark
  & \multicolumn{1}{c|}\cmark
  & \multicolumn{1}{c|}\xmark
  & \multicolumn{1}{c|}\xmark
  & \multicolumn{1}{c|}\cmark
  \\ \hline
  \begin{tabular}{@{}l@{}} Energy-Aware\end{tabular}
  & \multicolumn{1}{c|}\xmark
  & \multicolumn{1}{c|}\cmark
  & \multicolumn{1}{c|}\cmark
  & \multicolumn{1}{c|}\cmark
  & \multicolumn{1}{c|}\xmark
  & \multicolumn{1}{c|}\cmark
  & \multicolumn{1}{c|}\cmark
  & \multicolumn{1}{c|}\cmark
  \\ \hline
  \begin{tabular}{@{}l@{}} No-Offloading\end{tabular}
  & \multicolumn{1}{c|}\xmark
  & \multicolumn{1}{c|}\cmark
  & \multicolumn{1}{c|}\cmark
  & \multicolumn{1}{c|}\xmark
  & \multicolumn{1}{c|}\cmark
  & \multicolumn{1}{c|}\cmark
  & \multicolumn{1}{c|}\cmark
  & \multicolumn{1}{c|}\cmark
  \\ \hline
  \begin{tabular}{@{}l@{}} Continuous\end{tabular}
  & \multicolumn{1}{c|}\cmark
  & \multicolumn{1}{c|}\cmark
  & \multicolumn{1}{c|}\xmark
  & \multicolumn{1}{c|}\cmark
  & \multicolumn{1}{c|}\xmark
  & \multicolumn{1}{c|}\xmark
  & \multicolumn{1}{c|}\xmark
  & \multicolumn{1}{c|}\cmark
  \\ \hline
\end{tabular}
\caption{Comparison of the features offered by related works.}
\label{table:related_work}

\vspace{-3em}
\end{table}
\normalsize

The related work could be categorized as follows:

\textbf{Continous Detection:}
Glimpse~\cite{glimpse}, RoaD-RuNNer~\cite{roadrunner}, and FlexPatch~\cite{flexpatch} propose using an edge-server computational setup to decrease latency and reduce energy 
by using techniques such as edge level object tracking, edge/server model partitioning and selective tracking, respecitvely. 
However, such approaches rely on stable connections to servers, and none of them consider the use of multiple accelerators or multiple DNN models.
Marlin~\cite{marlin}, and AvaVP~\cite{adavp} 
are studies which aim to reduce the energy usage onboard a mobile device doing OD. 
Marlin~\cite{marlin} proposes an approach where, instead of running the DNN every frame, the system alternates between a tracking algorithm and DNN.
AvaVP~\cite{adavp} extends Marlin~\cite{marlin} by varying the input size of the DNN and skipping frames to adjust the \textit{e-a-l} trade-off during runtime.better latency per frame compared to other onboard-only schemes.
Neither considers the benefit of employing multiple DNN models or utilizing accelerators other than a GPU. \looseness=-1

\textbf{DNN Inference on Multi-Accelerator Systems:}
Inference on multi-accelerator systems is an active research area with several recent studies being published~\cite{axonn,herald,neulens,band, deepmon, codl, nestdnn}.
Herald~\cite{herald} and AxoNN~\cite{axonn} generate optimized schedules for performing inference with a single DNN on a system using a layer-by-layer mapping scheme. 
NeuLens~\cite{neulens} and Band~\cite{band} reduce DNN inference time by splitting a DNN into subgraphs for processing. 
Deepmon~\cite{deepmon} and CoDL~\cite{codl} optimize latency on mobile execution by scheduling layers on both CPU and GPU. 
While these solutions consider latency and energy constraints, neither of them  considers using multiple, different types of DNN models for OD in a context-aware manner. 

\textbf{Multi Model Detection:}
Fast UQ~\cite{nvidia-pose} uses different types of DNN architectures to extract more accurate poses in 3D space. They identify that combining multiple DNNs with domain specific metrics could lead to significant accuracy improvements.\looseness=-1

A feature comparison between the most relevant related work and our approach can be found in Table~\ref{table:related_work}. \sysname{} is uniquely able to handle context-aware, continuous multi-model (\ie{} multi-DNN) OD across multiple accelerators while satisfying energy, latency and accuracy constraints.

%% file: 03-methodology.tex
\section{Methodology}\label{chap:method}
\vspace{0.3em}

\sysname{} is composed of three primary components: (a) We first \textit{characterize} the target set of ODMs by determining their core traits for each available accelerator and build a \textit{confidence graph} to enable fast accuracy prediction at runtime. 
(b) Then, \textit{our multi-model, multi-accelerator scheduler} is responsible for determining the current best ODM to run, for each incoming frame. The scheduler uses the traits identified in the characterization process and the confidence graph, and selects the model for the next frame that most effectively meets accuracy, energy and latency constraints. 
(c) Finally, our scheduler employs a \textit{dynamic model loader} that utilizes characterization data and runtime memory footprint measurements to determine where to allocate models and, if necessary, which models to deallocate. 

\subsection{\textbf{Object Detection Model Characterization}}\label{section:model_characterization}

The latency, energy consumption, and detection accuracy of OD in an autonomous system depend on traits varying by the ODM, frame-context, and target accelerators. To select the best model for each frame encountered at runtime, we need to continously predict the accuracy of the ODMs at hand. However, this is not trivial unless each ODM is run everytime the frame-context changes, which is a very costly approach. To address this problem, we build a two-step approach where we first (1)  perform an offline characterization of ODMs into a discrete set of traits and  then, (2) using these traits, we construct a confidence graph
which will later be used by the runtime to predict the accuracy of a model. 
Notably, our approach for model characterization and graph construction is generic in nature. It relies solely on a testing or validation subset of the dataset used for training the models. \looseness=-1

\textbf{ODM Trait Identification:}
For each ODM, we collect the following traits:
    \textbf{(i)~Accuracy} is characterized by running the ODM on the testing data set and, for each frame, computing the intersection over union (IoU) between the bounding boxes of the detected object and labeled ground truth.
    \textbf{(ii)~Confidence score} 
    represents an internal accuracy assessment of the underlying DNN and can be used for quickly acquiring a performance estimate. However, these scores can be influenced by over-fitting and sometimes they are `over-confident';
    therefore, they are not consistent across different ODM architectures.
    \textbf{(iii)~Latency} of an ODM is found by measuring the execution time of the model on each target accelerator. 
    \textbf{(iv)~Energy} is characterized by measuring the $time \times power\_draw$ across all power rails during execution with a given model. 
    \textbf{(v)~Model loading cost} needs to be accounted for as part of multi-model execution overhead. This cost includes the memory footprint, time to load the model, and energy draw during this time. \looseness=-1

\textbf{Confidence Graph Creation:}
For a given frame, the confidence scores reported from multiple types of ODMs vary, while versions of the same ODM produce similar scores. Correlating the scores between different types is essential for accurate predictions, since we cannot run each model type on each frame due to energy, latency, and memory constraints.

To quickly predict the accuracy of ODMs on the fly, we construct a novel \textit{confidence graph} (CG) as follows:\looseness=-1

\begin{enumerate}[wide=\parindent]
  \renewcommand{\labelitemi}{$\Rightarrow$}
 \item Each node represents a discrete confidence score range of an ODM and its expected accuracy (\eg{} we create a node, \texttt{YoloV7-(0.5-0.6)}, to represent YoloV7's performance in the confidence score range of 0.5 to 0.6). 
 \item To find edges, we run each ODM on a testing/validation dataset. For each image in the dataset, we locate the nodes for each ODM's confidence score range on that image and create edges between them. For example, for a given image, if YoloV7 resulted in a conf. score of 0.53 and MobileNet results in 0.42, we create an edge between the two corresponding nodes: \texttt{YoloV7-(0.5-0.6)} and \texttt{MobileNet-(0.4-0.5)}. If an edge already exists, we increment its weight by 1. \looseness=-1
 \item We then normalize all edges to have a weight $w$ between 0 and 1 and invert weights so that two highly connected nodes will have a lower cost when traversed. The normalization is performed within the edges directly connecting a single node, such that global maximums will not take over. \looseness=-1
 \item Next we run a breadth first search starting at all nodes in the graph, to get the set of neighbor nodes which have a distance less than or equal to a given distance threshold. \looseness=-1
 \item Since a given set of neighbors can contain multiple nodes representing the same model, said nodes are consolidated by taking a weighted average of each node's expected accuracy by the distance to traverse to said node.
 \item We then store all results from the CG in a map where each node is a key to the accuracy predictions of its neighbors. \looseness=-1
\end{enumerate}

In summary, our proposed CG structure generates a map that transforms the confidence score of a single ODM into accuracy predictions for all ODMs. Instead of relying on costly classifiers, an ensemble, or less expensive predictors employed by similar works~\cite{marlin, adavp}, we can execute a map lookup at runtime.

\subsection{\textbf{\sysname{} Scheduler}}

The \sysname{} scheduler is designed to perform the decision-making process at runtime, leveraging contextual information from both model runtime behavior and the input data stream (\ie{} frames). 
Our scheduler can make fast decisions with minimal computational overhead by utilizing the \textit{CG} and the analysis of frames using computationally efficient metrics. As a result, the scheduler maintains an overhead of less than 2 milliseconds per frame. The scheduler operation can be delineated into two fundamental components: context detection, and a heuristic algorithm for selecting appropriate models.

\textbf{Context Detection:}
\sysname{} scheduler relies on
input frames to detect the changes in the context, so that the proper ODM could be employed. 
While the \textit{CG} lets us rapidly predict the accuracy of an ODM by deriving information from confidence scores reported by DNNs,
these  scores are intrinsically linked to model error~\cite{uncertainty-framework} and may not be reliable when the input data is further outside the scope of the training data than the testing set. 
The frame context also plays a crucial role in predicting the accuracy of an ODM ~\cite{direct-uncertainty}. Extracting a comprehensive range of context from these images is computationally expensive and not viable for real-time processing on edge devices. Instead, the \sysname{} scheduler assesses frame similarity using the normalized cross-correlation (NCC) between consecutive bounding box results and image frames: 
\begin{equation}
\small
\label{eq:ncc}
\text{NCC}(p,c) = \frac{\sum{(p-mean(p))(c-mean(c))}}{(\sqrt{\sum{(c-mean(c))}^2} \times \sqrt{\sum{(p-mean(p))}^2}}
\end{equation}
where $p$ and $c$ are grayscale images of the same size representing the previous and current frames in an input stream. By employing NCC, the scheduler can identify when the input stream has changed significantly, prompting re-scheduling of the current ODM. This can aid in identifying when the current ODM may incorrectly continue to report high confidence scores despite objects not being present or easily detectable.

\textbf{Scheduling Heuristic:}
Algorithm \ref{alg:sched} describes the \sysname{} scheduler which utilizes a heuristic to assign weights to schedulable models. 
The scheduler takes the current ODM $m$, conf. score $c$, input frame $i$ and bounding box $b$ as input and returns the highest scoring model as output. The energy and latency characteristics of available models are pre-determined, normalized to a 0 to 1 range, and inverted for bigger-is-better performance indication (lines 6, 7). At runtime, the scheduler invokes the confidence graph to estimate model accuracies for the most recent frame (line 9). The scheduler averages accuracy predictions for all models and aggregates those meeting the desired accuracy threshold (lines 11-15). In the absence of models meeting the threshold, all available models are considered valid (lines 16-17). Post model selection, weights are assigned based on user-defined parameters, and the best model is outputted (lines 19-24). As energy and latency are converted to bigger-is-better metrics, maximum search of candidate models suffices for optimal selection.\looseness=-1

The image similarity score is computed as the minimum of the NCC between the last two images and the NCC across the last two bounding box detections. This score determines whether the scheduler should initiate the selection of a new model/accelerator pair. The metric is then multiplied by the current model confidence to facilitate scheduling during periods when the existing ODM may exhibit unstable detections. By reserving the scheduling of new models for instances characterized by rapid changes in the overall image context, bounding box, or a reduction in model confidence, the overall cost incurred by model swapping can be minimized.

\begin{figure}

\vspace{-0.7em}

\begin{algorithm}[H]
\begin{algorithmic}[1]
\small
\Procedure{\sysname{} Schedule}{$m, c, i, b$}
    \State $s = \text{min}(\text{NCC}(\text{lastImage},i),\text{NCC}(\text{lastBbox},b))$
    \If{$ s \times c \geq accuracyThreshold$}
        \State \Return $\text{m}$
    \EndIf
    \State $E = \text{scheduler.energy}$ \Comment{$0 \rightarrow 1$ model energy}
    \State $L = \text{scheduler.latency}$ \Comment{$0 \rightarrow 1$ model latency}
    \State $W = \text{scheduler.weights}$ \Comment{Tuned knobs}
    \State $C = \text{graphPredict}(m, c)$ \Comment{set of (name, acc, dist)}
    \State $R, scores = \text{map}(), \text{map}()$
    \For{$(n, a, d) \in C$}
        \State $\text{a.Buffer}.append(a)$
        \State $R[n] = \text{average}(\text{a.Buffer})$
    \EndFor
    \State $V = \set{n \mid n \in R, n \geq accuracyThreshold}$
    \If{$\text{length}(V) == 0$}
        \State $V = R$
    \EndIf
    \For{$n \in R.\text{keys}()$}
        \State $s = R[n] * W[0] + E[n] * W[1] + L[n] * W[2]$
        \State $scores[n] = s$
    \EndFor
    \State \Return $\text{max}(scores)$
\EndProcedure
\end{algorithmic}
\caption{Model Scheduling}\label{alg:sched}
\end{algorithm}

\vspace{-2.5em}

\end{figure}

\subsection{\textbf{Dynamic Model Loader}}

Every model which can be executed has profiling information available about the process of loading the model into memory. When there is a scheduling decision and a new model is requested to be loaded into memory, the dynamic model loader (DML) will query the system's available memory. The DML will attempt to occupy the entire memory with ODMs, if it is able to. This reduces the costly switching between ODMs and improves the performance by enabling quicker model swapping. The DML is able to differentiate between accelerators and will allocate to them separately. When replacing models the DML will replace the model which was least recently requested. Since accelerators do not all share the same memory, the DML needs to have the knowledge about whether an accelerator can execute a specific ODM. \looseness=-1

%% file: 04-experiments.tex
\section{Experimental Setup}\label{chap:experiments}

\textbf{Hardware and Accelerators:}\label{section:hardware}
We performed our experiments on Nvidia Xavier NX SoC, a commonly used platform for aerial autonomous vehicles, and a Luxonis OAK-D Lite, a stereo camera with DNN execution capability. The platform includes a CPU, GPU, 2 DLAs, and an OAK-D for DNN execution.
Due to model layer incompatibility, limitations on model size, and support constraints in libraries, the DLA and OAK-D do not support some layers and models we use in our experiments.\looseness=-1

\textbf{Dataset:}\label{section:dataset}
We train the ODMs with a dataset focused on the detection of unmanned autonomous vehicles (UAVs)~\cite{drone-data}, which comprises a training set of ~50,000 images and a validation set of ~2,500 images. Each image contains at most a single UAV, hence all implementation and evaluation are within the context of a single-class, single-object detection problem.

\textbf{Model Training:}\label{section:model_training}
YoloV7~\cite{yolov7} based models were trained using the training scripts and pipelines provided by the authors. All YoloV7 models use an IoU threshold of 0.5 and a confidence threshold of 0.35 in non-maximum-suppression. MobileNetV1
, MobileNetV2
, and Resnet50
 were trained with the single-shot detector
 methodology. These models were trained using the Tensorflow object detection API.
 All models use an input size of 640x640 unless otherwise stated.

\textbf{Method Evaluation \& Comparison:}\label{section:model_evaluation}
Comprehensive assessments were conducted across various standalone ODMs, considering their compatibility with distinct accelerators. 
Our analysis includes a comparative analysis with Marlin \cite{marlin}, recognized for its energy-efficient executions on mobile devices. 

To test all methods' ability to handle real-life challenges we created a custom evaluation dataset consisting of six scenarios as a set of videos, each comprising between 500 to 2,500 frames. The videos have two indoor and four outdoor scenarios where the targeted UAV is at different distances from the camera with varying backgrounds and positioning in the frame. 

All models on the GPU/DLA were executed using TensorRT.
OpenVINO
was used to compile models for the OAK-D.
All GPU layers for models were executed in FP32 due to severe accuracy degradation during quantization with TensorRT 
for YoloV7 models.
To establish a performance ceiling, an Oracle methodology was created. This Oracle identifies all models surpassing a 0.5 intersection-over-union (IoU) threshold, subsequently selecting the one that optimizes the targeted metric. In cases where no models meet the IoU criterion, selection is solely based on metric optimization. Since the Oracle methods represents a maximum performance, it assumes that all models are loaded into memory and thus have no cost to switch. The choice of a 0.5 IoU threshold aligns with the common practice of using a 0.5 threshold as the minimum when evaluating object detection models \cite{yolov7
}. Additionally, we define the metric \textit{success rate} as the percentage of frames which have an IoU $\geq$ 0.5. Since the testing videos contain a single UAV, the average IoU effectively captures all relevant accuracy information.

%% file: 05-results.tex
\section{Evaluation}\label{chap:results}

\subsection{Main results}

\begin{figure}[t]
\begin{center}
    \includegraphics[width=\columnwidth]{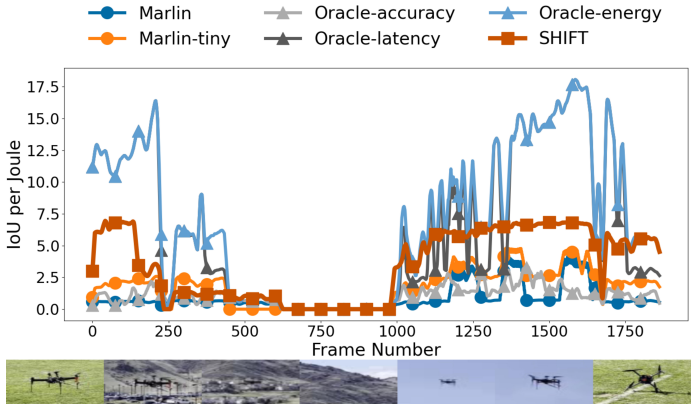}
    \caption{Scenario 1: Drone navigates across multiple backgrounds at \textit{varying} distances from the camera.}
    \label{fig:scenario1}
\end{center}
\vspace{-2em}
\end{figure}

\begin{figure}[t]
\begin{center}
    \includegraphics[width=\columnwidth]{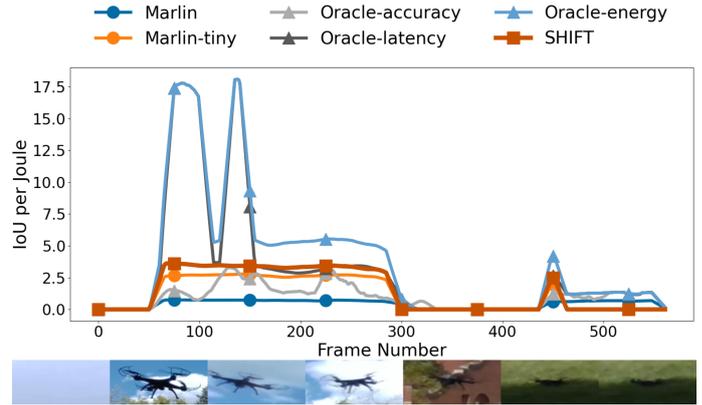}
    \caption{Scenario 2: Drone navigates across multiple backgrounds at a \textit{fixed} distance.}
    \label{fig:scenario2}
\end{center}
\vspace{-3.08em}
\end{figure}

The detailed traits from the characterization of all ODMs we employ are given in Table ~\ref{table:models}. Table ~\ref{table:methods} presents the overall results of our experiments, by giving a comparison between average accuracy, latency and energy obtained by Marlin~\cite{marlin}, \sysname{} and Oracle executions. Additionally, Figures~\ref{fig:scenario1} and \ref{fig:scenario2} illustrate the accompanying timelines and frame context changes for two of the videos contributing to Table~\ref{table:methods}. 

Overall averages given in Table~\ref{table:methods} show that \sysname{} consistently achieves a success rate surpassing all single-model executions except the top-performing YoloV7 model. \textit{Uniquely, \sysname{} manages to uphold energy and latency efficiency superior to Marlin and all single-model runs while consistently maintaining desired accuracy ranges.} Notably, \sysname{} outperforms the state-of-the-art method Marlin irrespective of the underlying ODM which Marlin utilizes. 

In the first scenario, which is given in Figure~\ref{fig:scenario1}, the UAV executes maneuvers across intricate backgrounds distant from the camera before returning. A proficient dynamic system must adeptly identify crucial context changes, as evidenced by sharp fluctuations in model efficiency values. \sysname{} successfully recognizes all context changes, implementing transitions to more resource-intensive ODMs at frame markers  $\sim500$ and $\sim1100$. Simultaneously, it strategically conserves resources by transitioning at frames $\sim50$ and $\sim1650$. The unique capability of \sysname{} to augment resource usage during challenging or simple inputs contributes to its superiority over solutions like Marlin. Importantly, \sysname{} can conservatively allocate resources during periods without valid detections.

\begin{table}[b]
\vspace{-2em}
\setlength\tabcolsep{2.5pt}
    \begin{tabularx}{\columnwidth}{@{} l | *{7}{L} @{}}
          \toprule
          Methodology & IoU & Time (s) & Energy (J) & Success Rate & Non-GPU & Model Swaps & Pairs Used\\
          \midrule
          Marlin & 0.614 & 0.132 & 1.201 & 74.0\% & 0\% & 0 & 1 \\
          Marlin Tiny & 0.529 & 0.036 & 0.33 & 64.0\% & 0\% & 0 & 1 \\
          \textbf{\sysname{}} & 0.598 & 0.047 & 0.262 & 72.2\% & 68.7\% & 42 & 4.3\\
          Oracle E & 0.535 & 0.025 & 0.144 & 76.0\% & 31.5\% & 94 & 6.7 \\
          Oracle A & 0.657 & 0.108 & 1.423 & 76.0\% & 44.9\% & 409 & 12.3 \\
          Oracle L & 0.522 & 0.025 & 0.169 & 76.0\% & 11.3\% & 112 & 6.8 \\
          \bottomrule
    \end{tabularx}
    \caption{Average runtime performance of continuous object detection with \sysname{}. \sysname{} parameters: goal accuracy 0.25, momentum 30, distance threshold 0.5, knobs: accuracy 1.0, and energy/latency 0.5. Goal accuracy reduced from 0.5, based on observation from Figure~\ref{fig:sensitivity}. Pairs are model accelerator pairs, a total of 18 combinations were possible. Includes overhead for \sysname{} and Marlin methods.}
    \vspace{-0.2em}
    \label{table:methods}
\end{table}
\begin{table*}
    \setlength\tabcolsep{3pt}
    \begin{tabular*}{\textwidth}{@{\extracolsep{\fill}}lccccccccccc}
        \toprule
        \multirow{1}{*}{Model Name} & \multicolumn{2}{c}{Accuracy} & \multicolumn{3}{c}{Avg. Time (s)} & \multicolumn{3}{c}{Avg. Energy (Joules)} & \multicolumn{3}{c}{Avg. Power Draw (W)} \\
        & Avg. IoU & Success Rate & GPU & GPU/DLA & OAK-D & GPU & GPU/DLA & OAK-D & GPU & GPU/DLA & OAK-D \\
        \midrule
        YoloV7-E6E & 0.564 & 65.8\% & 0.255 & 0.221 & - & 3.947 & 1.228 & - & 15.48 & 5.56 & - \\
        YoloV7-X & 0.593 & 71.1\% & 0.222 & 0.195 & - & 3.586 & 1.088 & - & 16.15 & 5.57 & - \\
        YoloV7 & 0.618 & 74.1\% & 0.130 & 0.118 & 0.894 & 1.968 & 0.656 & 1.391 & 15.14 & 5.56 & 1.56 \\
        YoloV7-Tiny & 0.533 & 64.0\% & 0.025 & 0.024 & 0.107 & 0.280 & 0.134 & 0.206 & 11.2 & 5.58 & 1.93 \\
        SSD Resnet50 & 0.480 & 58.9\% & 0.151 & 0.138 & - & 2.504 & 0.816 & - & 16.58 & 5.91 & - \\
        SSD MobilenetV1 & 0.452 & 55.4\% & 0.094 & 0.092 & - & 1.519 & 0.561 & - & 16.16 & 6.10 & -\\
        SSD MobilenetV2 & 0.401 & 51.3\% & 0.023 & 0.058 & - & 0.248 & 0.307 & - & 10.78 & 5.29 & - \\
        SSD MobilenetV2 320x320 & 0.304 & 36.2\% & 0.009 & 0.023 & - & 0.046 & 0.100 & - & 5.11 & 4.35 & - \\
        \bottomrule
    \end{tabular*}
    \caption{Collected accuracy and performance traits of all models.}\label{table:models}
\end{table*}

In the second scenario given in Figure~\ref{fig:scenario2}, the UAV moves horizontally across simpler backgrounds while gradually moving across the camera's perspective. Analogous to the first scenario, we observe abrupt drops in ODM accuracy, and thus efficiency, when the UAV is between backgrounds. Notably, \sysname{} successfully identifies when the UAV enters the camera view, prompting model swaps to enhance efficiency. The discernable delay in \sysname{}'s response compared to Marlin and the Oracles is attributed to its reactionary model swapping. It is noted that \sysname{} did not detect the UAV beyond frame $\sim450$ due to confidence scores of the current ODM indicating no UAV. 

From Table~\ref{table:methods}, \sysname{} utilizes fewer ODM-accelerator pairs than the Oracle methods while still maintaining higher efficiency than previous state-of-the-art Marlin~\cite{marlin}. Despite higher utilization of heterogeneous resources compared to the Oracles, \sysname{} loads less models into memory and uses fewer swaps between ODMs or accelerators. For scenario 1 (pictured in Figure~\ref{fig:scenario1}), \sysname{} only loaded ODMs which were smaller than YoloV7; hence showcasing \sysname{}'s preference for efficient inference via the tunable weights.

Overall, we observe that \sysname{} is able to successfully optimize across energy and latency while swapping ODMs during runtime based on contextual information from the input stream in all real-world scenarios evaluated.

\subsection{Sensitivity analysis}

\begin{figure}[b]
\begin{center}
    \vspace{-1em}
    \includegraphics[scale=0.27]{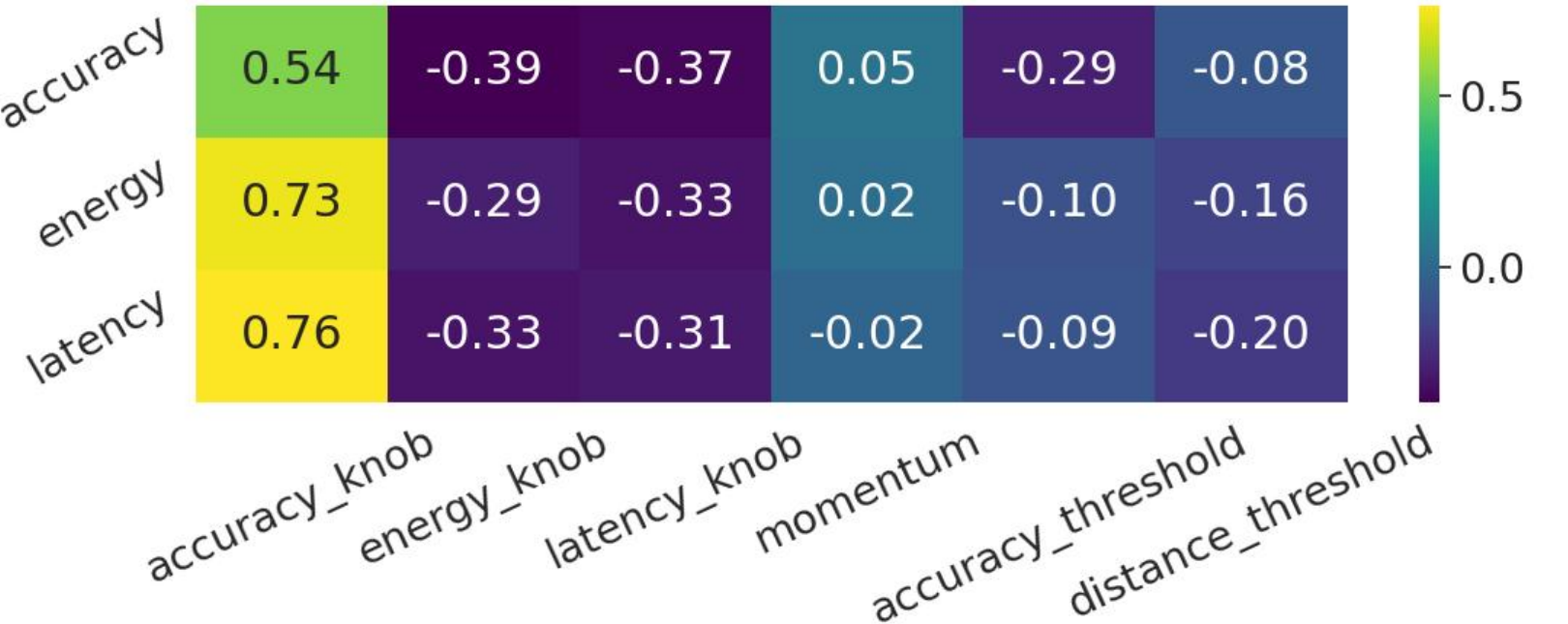} 
    \caption{Sensitivity analysis of the \sysname{} parameters against the mean accuracy, energy, and latency values.}
    \label{fig:sensitivity}
\end{center}
\end{figure}

We performed a sensitivity analysis on \sysname{} to ascertain the robustness of system performance against variations in input parameters. 
A total of 1860 parameter configurations were tested. The outcomes of this analysis are depicted in Figure~\ref{fig:sensitivity}. The analysis indicates that the system's performance conforms to expectations with respect to all knob parameters. By increasing the value of the energy or latency knob, we observe a negative correlation with the actual ODM's energy and latency as expected. Conversely, the accuracy knob has a positive correlation since more expensive ODMs are more accurate. The accuracy threshold parameter inversely affects all primary metrics; this is because when \sysname{} fails to find any models meeting the goal accuracy it defaults to optimization based on the knob settings alone. We observe that ODM accuracy is underestimated and lowering the accuracy goal for runtime improves efficiency. The momentum parameter, indicative of the number of frames over which to average the predicted accuracy of a given ODM, exhibits minor correlation with the performance metrics, suggesting that frame-to-frame results are inherently stable. Additionally, the distance threshold parameter has a distinct correlation with reducing the average ODM latency, due to more ODMs being considered at runtime.

%% file: 06-conclusion.tex
\section{Conclusion}\label{chap:conclusion}
\vspace{0.3em}

We introduce \sysname{}, capable of dynamically switching between heterogeneous DNNs and target hardware based on continuous input stream context. \sysname{} facilitates multi-model swaps for accuracy, energy, and latency trade-offs, leveraging multiple accelerators. Experimental results demonstrate \sysname{}'s effectiveness, showing significant improvements. Compared to a state-of-the-art ODM on GPU, \sysname{} achieves up to a \textbf{2.8x} reduction in latency and a \textbf{7.5x} decrease in energy consumption, with only a modest 0.97x reduction in successful frames and 0.97x reduction in average IoU. Notably, \sysname{} maintains performance without inter-frame object tracking or skipping input frames.

%% file: paper.bbl
\begin{thebibliography}{10}
\providecommand{\url}[1]{#1}
\csname url@samestyle\endcsname
\providecommand{\newblock}{\relax}
\providecommand{\bibinfo}[2]{#2}
\providecommand{\BIBentrySTDinterwordspacing}{\spaceskip=0pt\relax}
\providecommand{\BIBentryALTinterwordstretchfactor}{4}
\providecommand{\BIBentryALTinterwordspacing}{\spaceskip=\fontdimen2\font plus
\BIBentryALTinterwordstretchfactor\fontdimen3\font minus \fontdimen4\font\relax}
\providecommand{\BIBforeignlanguage}[2]{{%
\expandafter\ifx\csname l@#1\endcsname\relax
\typeout{** WARNING: IEEEtran.bst: No hyphenation pattern has been}%
\typeout{** loaded for the language `#1'. Using the pattern for}%
\typeout{** the default language instead.}%
\else
\language=\csname l@#1\endcsname
\fi
#2}}
\providecommand{\BIBdecl}{\relax}
\BIBdecl

\bibitem{yolov7}
C.-Y. Wang, A.~Bochkovskiy, and H.-Y.~M. Liao, ``Yolov7: Trainable bag-of-freebies sets new state-of-the-art for real-time object detectors,'' in \emph{CVPR'23}.

\bibitem{glimpse}
T.~Y.-H. Chen, L.~Ravindranath, S.~Deng, and et~al., ``Glimpse: Continuous, real-time object recognition on mobile devices,'' in \emph{SenSys'15}.

\bibitem{flexpatch}
K.~Yang, J.~Yi, K.~Lee, and Y.~Lee, ``Flexpatch: Fast and accurate object detection for on-device high-resolution live video analytics,'' in \emph{INFOCOM'22}.

\bibitem{adavp}
M.~Liu, X.~Ding, and W.~Du, ``Continuous, real-time object detection on mobile devices without offloading,'' in \emph{ICDCS'20}.

\bibitem{marlin}
K.~Apicharttrisorn, X.~Ran, J.~Chen, and et~al., ``Frugal following: Power thrifty object detection and tracking for mobile augmented reality,'' in \emph{SenSys'19}.

\bibitem{framehopper}
M.~Adnan~Arefeen, S.~Tabassum~Nimi, and M.~Yusuf Sarwar~Uddin, ``Framehopper: Selective processing of video frames in detection-driven real-time video analytics,'' in \emph{DCOSS'22}.

\bibitem{axonn}
I.~Dagli, A.~Cieslewicz, J.~McClurg, and M.~E. Belviranli, ``Axonn: Energy-aware execution of neural network inference on multi-accelerator heterogeneous socs,'' in \emph{DAC'22}.

\bibitem{gholami2021survey}
A.~Gholami, S.~Kim, Z.~Dong, and et~al., ``A survey of quantization methods for efficient neural network inference,'' \emph{Low-Power Computer Vision}, 2021.

\bibitem{roadrunner}
A.~K. Kakolyris, M.~Katsaragakis, D.~Masouros, and D.~Soudris, ``Road-runner: Collaborative dnn partitioning and offloading on heterogeneous edge systems,'' in \emph{DATE'23}.

\bibitem{nvidia-pose}
G.~Shi, Y.~Zhu, J.~Tremblay, and et~al., ``Fast uncertainty quantification for deep object pose estimation,'' in \emph{ICRA'21}.

\bibitem{herald}
H.~Kwon, L.~Lai, M.~Pellauer, and et~al., ``Heterogeneous dataflow accelerators for multi-dnn workloads,'' in \emph{HPCA'21}.

\bibitem{neulens}
X.~Hou, Y.~Guan, and T.~Han, ``Neulens: Spatial-based dynamic acceleration of convolutional neural networks on edge,'' in \emph{MobiCom '22}.

\bibitem{band}
J.~S. Jeong, J.~Lee, D.~Kim, C.~Jeon, C.~Jeong, Y.~Lee, and B.-G. Chun, ``Band: Coordinated multi-dnn inference on heterogeneous mobile processors,'' in \emph{MobiSys'22}.

\bibitem{deepmon}
L.~N. Huynh, Y.~Lee, and R.~K. Balan, ``Deepmon: Mobile gpu-based deep learning framework for continuous vision applications,'' in \emph{MobiSys'17}.

\bibitem{codl}
F.~Jia, D.~Zhang, T.~Cao, S.~Jiang, Y.~Liu, J.~Ren, and Y.~Zhang, ``Codl: Efficient cpu-gpu co-execution for deep learning inference on mobile devices,'' in \emph{MobiSys '22}.

\bibitem{nestdnn}
B.~Fang, X.~Zeng, and M.~Zhang, ``Nestdnn: Resource-aware multi-tenant on-device deep learning for continuous mobile vision,'' in \emph{MobiCom'18}.

\bibitem{uncertainty-framework}
A.~Loquercio, M.~Segu, and D.~Scaramuzza, ``A general framework for uncertainty estimation in deep learning,'' \emph{IEEE Robotics and Automation Letters}, 2020.

\bibitem{direct-uncertainty}
M.~Raghu, K.~Blumer, R.~Sayres, Z.~Obermeyer, R.~Kleinberg, S.~Mullainathan, and J.~Kleinberg, ``Direct uncertainty prediction for medical second opinions,'' in \emph{ICML'19}.

\bibitem{drone-data}
M.~L. Pawelczyk and M.~Wojtyra, ``Real world object detection dataset for quadcopter unmanned aerial vehicle detection,'' \emph{IEEE Access}, 2020.

\end{thebibliography}
